%% file: acl_latex.tex
\documentclass[11pt]{article}

\usepackage[preprint]{acl}

\usepackage{times}
\usepackage{latexsym}

\usepackage[T1]{fontenc}
\usepackage[utf8]{inputenc}

\usepackage{microtype}

\usepackage{inconsolata}

\usepackage{graphicx}

\usepackage{amsmath}
\usepackage{xspace}
\usepackage{enumitem}

\def\method{\textsc{CiteAnchor}\xspace}
\def\dataset{\textsc{CiteResolve}\xspace}

\title{\method: Source-Aware Grounding for \\
Hallucinated Citation Detection and Resolution
}

\author{
  \textbf{Chenyi Tong\textsuperscript{1}}
  \quad
  \textbf{Ping Nie\textsuperscript{2}}
  \quad
  \textbf{Yu Zhang\textsuperscript{3}}
  \\
  \textsuperscript{1}University of Wisconsin--Madison
  \quad
  \textsuperscript{2}University of Waterloo
  \quad
  \textsuperscript{3}Texas A\&M University
  \\
  \texttt{ctong29@wisc.edu}
  \quad
  \texttt{yuzhang@tamu.edu}
}

\usepackage{tikz}
\usetikzlibrary{positioning,arrows.meta,calc,fit}
\usepackage{pifont}
\usepackage{amssymb}

\usepackage{booktabs}
\usepackage{colortbl}
\usepackage{tabularx}
\definecolor{refbg}{HTML}{E8F2F8}
\definecolor{backbonebg}{HTML}{D9F0EA}

\usepackage[skins,breakable]{tcolorbox}  % instruction templates in Appendix~\ref{app:prompts}
\definecolor{promptframe}{HTML}{3333B2}
\definecolor{promptback}{HTML}{F3F7FE}
\definecolor{promptstr}{HTML}{B22222}
\newtcolorbox{promptbox}[1]{
  enhanced,
  breakable,
  colback=promptback,
  colframe=promptframe,
  colbacktitle=promptframe,
  coltitle=white,
  fonttitle=\bfseries,
  title={#1},
  boxrule=0.7pt,
  arc=2pt,
  left=5pt,
  right=5pt,
  top=4pt,
  bottom=4pt,
  fontupper=\small
}

\newcommand{\pstr}[1]{\textcolor{promptstr}{#1}}   % JSON strings inside a prompt box

\usepackage{fontawesome5}

\newcommand{\reported}{\textsuperscript{\#}}

\addtocontents{toc}{\protect\setcounter{tocdepth}{-1}}

\begin{document}
\maketitle

\input{0_abstract}
\input{1_intro}
\input{2_related}

\input{3_benchmark}

\input{4_method}
\input{5_experiment}
\input{6_analysis}

\input{7_conclusion}



\bibliography{custom}

\input{8_appendix}

\end{document}

%% file: 0_abstract.tex
\begin{abstract}
Hallucinated citations make it harder to trace sources and verify the evidence behind scientific claims.
Detecting them is difficult because bibliographic records can be incomplete or differ across source types and publication versions.
A plausible match alone neither validates the citation nor establishes its underlying source.
We introduce \dataset, a benchmark of 423 real-world hallucinated citations annotated for source recoverability.
We further propose \method, a source-aware grounding framework that adapts retrieval and field-level verification to each source type and separately determines citation validity and whether a unique source can be recovered.
On an existing detection benchmark, \method achieves 0.841 F1 with GPT-5.6 Luna and 0.784 with an open-weight backbone, outperforming the baseline with the same backbones by 7.9 and 36.8 points.
On \dataset, it achieves 74.7\% and 70.7\% end-to-end task success, outperforming the corresponding web search baselines by 16.3 and 65.3 percentage points.
Our code and data are available at \url{https://github.com/oldwizard7/CiteANCHOR_code}.
\end{abstract}

%% file: 1_intro.tex
\section{Introduction}

% Large language models are increasingly being used to support scientific
% research, from literature synthesis~\citep{asai2026synthesizing}
% to agent-assisted scientific discovery~\citep{
% boiko2023autonomous,roohani2025biodiscoveryagent,tong2026crystalstar}.
% As AI systems participate more deeply in scientific writing and reasoning,
% the reliability of their references becomes increasingly important.
% Hallucinated citations remain a persistent failure mode, ranging from
% nonexistent references to real works described with incorrect
% bibliographic information~\citep{sakai2026hallucitation,xu2026ghostcite}.
% Reliable and scalable citation verification is therefore becoming an
% important component of AI-assisted research.

Large language models (LLMs) are increasingly integrated into the practice of scientific paper writing~\citep{liang2025quantifying,kusumegi2026scientific}.
When these models generate references, however, they can invent papers or mix correct and incorrect details about real ones.
Such \emph{hallucinated citations} have already been found in published conference papers and across the scientific literature~\citep{sakai2026hallucitation,xu2026ghostcite,zhao2026wild}.
Their harms are evident: they can lend credibility to unsupported claims, spread misinformation through the scientific literature, contaminate the scientific knowledge base, and undermine decisions in high-stakes domains.

Automated tools now combine bibliographic lookup, web search, and LLM judgments to detect these errors at scale~\citep{sakai2026hallucitechecker,shi2026citeaudit,li2026source,khajavi2026citecheck}.
Automated reference checks have already been used at ICLR 2026\footnote{\url{https://blog.iclr.cc/2026/03/31/a-retrospective-on-the-iclr-2026-review-process/}} and NeurIPS 2026\footnote{\url{https://www.reddit.com/r/MachineLearning/comments/1w7t71d/neurips_2026_does_receiving_the_automatic/}}, with flagged references passed to human reviewers.
For such screening to be useful, a system must explain why a citation is flagged and provide source records that allow reviewers to check its decision.
Here, two challenges arise.

\smallskip
\noindent \textbf{Interpreting Heterogeneous Source Evidence.}
Papers, books, and web resources differ in how they can be retrieved and which bibliographic fields identify them.
Even for the same work, databases may omit different fields, while preprint and published versions may differ in year or venue~\citep{visser2021large,shi2026citeaudit}.
These differences make simple matching unreliable: a valid citation may omit a field present in a database record, while an invalid citation may match a record that omits the erroneous field.
Reliable detection therefore requires source-aware interpretation of the evidence, distinguishing genuine contradictions from missing information and legitimate publication variants.

\smallskip
\noindent \textbf{Separating Citation Errors from Source Identity.}
A citation with incorrect authors may still refer to an identifiable paper, whereas one with an entirely fabricated title may have no real source at all.
Both should be classified as hallucinated, but evidence that establishes an error does not necessarily establish a replacement.
Recovering the correct source makes the discrepancy explicit and gives reviewers evidence to inspect.
Selecting a merely plausible source, however, can introduce another false claim.
A reliable system must therefore distinguish whether a citation is wrong from whether its underlying work can be uniquely identified, and abstain from correction when that identity remains unresolved.

To evaluate this distinction, we introduce the \textbf{\dataset} benchmark for source resolution in real-world hallucinated citations.
Building on CiteAudit~\citep{shi2026citeaudit}, we annotate recoverability and provide verified corrected references for citations with a uniquely supported source.
\dataset evaluates correct recovery and appropriate abstention in an end-to-end setting, where systems receive \textit{only} the original citation and must determine whether it is hallucinated and, if so, identify a uniquely supported source or conclude that no unique source can be established.

We further propose \textbf{\method}, an agentic framework built on \emph{source-aware grounding}: interpreting retrieved records using source-specific criteria for citation verification and source resolution.
Its central mechanism is \emph{evidence decomposition}: checking individual bibliographic fields before making separate decisions about citation validity and source identity.
A chain-of-source strategy implements this decomposition through type-specific retrieval, deterministic bibliographic checks, and LLM reasoning for cases that these checks do not settle.
Retrieved evidence is shared across discovery, verification, and resolution, with a targeted follow-up search when a detected error still has no supported source.
This design grounds both hallucination labels and proposed corrections in verifiable records.

We evaluate \method with the proprietary GPT-5.6 Luna~\citep{openai2026gpt56luna} and the open-weight OpenResearcher-30B-A3B~\citep{li2026openresearcher}.
On real-world hallucinated citation detection, \method improves F1 over the CiteAudit baseline~\citep{shi2026citeaudit} with the same backbones by 7.9 and 36.8 points, respectively.
Even with the open-weight backbone, which supports local inference for confidential submissions, it exceeds the direct-prompting results reported by \citet{shi2026citeaudit} for Gemini-3-Pro~\citep{google2025gemini3pro}, GPT-5.2~\citep{openai2025gpt52}, and Claude-Sonnet-4.5~\citep{anthropic2025sonnet45}.
On our \dataset benchmark, overall resolution success reaches 74.7\% and 70.7\%, exceeding the corresponding Web Search baselines by 16.3 and 65.3 percentage points.
With the proprietary backbone, replacing our staged procedure with a single LLM judgment over the same retrieved evidence reduces detection F1 by 19.8 points, while removing source-type adaptation reduces it by 8.0 points.
These ablations support the importance of evidence handling beyond retrieval alone.
The full workflow also uses 70.7\% fewer commercial search API calls on average across the two backbones than CiteAudit, corresponding to an estimated saving of approximately \$25 in search credits per run over 3,356 citations.

Our contributions are threefold:
\begin{itemize}[nosep,leftmargin=*]
\item \textbf{Conceptually}, we extend hallucinated citation evaluation to source identity with the \dataset benchmark, measuring correct recovery and justified abstention against verified targets.
\item \textbf{Methodologically}, we introduce source-aware grounding through evidence decomposition, combining type-specific retrieval, explicit field checks, and separate validity and resolution decisions in our \method framework.
\item \textbf{Empirically}, we show stronger detection and overall source resolution across proprietary and open-weight backbones, supported by controlled ablations and substantially fewer commercial search API calls.
\end{itemize}

%% file: 2_related.tex
\section{Related Work}
\label{sec:related}

\noindent \textbf{Hallucinated References and Bibliographic Verification.}
Hallucinated references occur in both LLM-generated texts~\citep{walters2023fabrication} and published papers~\citep{sakai2026hallucitation,xu2026ghostcite,zhao2026wild}, while uneven database coverage and incomplete records complicate verification~\citep{visser2021large}.
CiteAudit~\citep{shi2026citeaudit} addresses this problem through multi-stage bibliographic lookup, web search, and evidence-based judgments.
CiteTracer~\citep{li2026source} combines a retrieval cascade with deterministic field checks and LLM judgments, distinguishing missing information from contradictions.
CiteCheck~\citep{khajavi2026citecheck} also identifies candidate publications before checking their fields and distinguishes references to identifiable works with corrupted metadata from those without credible matches, evaluating these labels on synthetic corruptions.
HALLMARK~\citep{reizinger2026hallmark} evaluates verification under metadata errors and highlights false positives as a deployment concern.
Our work builds on these retrieval and verification strategies through source-aware evidence decomposition, and extends evaluation to the correctness of recovered sources and abstention decisions against verified targets.

\smallskip
\noindent \textbf{Citation Resolution and Correction.}
In bibliometrics, reference matching consolidates noisy variants of the same work within a dataset through normalization and metadata reconciliation~\citep{aria2026multiphase}.
Scientific citation attribution starts from the surrounding text: CiteME~\citep{press2024citeme} evaluates recovery of author-cited papers, while CiteGuard~\citep{choi2026citeguard} also identifies alternative valid citations.
In retrieval-augmented generation, ALCE~\citep{gao2023enabling} evaluates citation support in generated answers, and CiteFix~\citep{maheshwari2025citefix} corrects citation assignments against retrieved documents.
Our setting starts from the bibliographic entry itself, without the surrounding claim: the system must check whether it is valid and determine whether its remaining information identifies a unique real source.

%% file: 3_benchmark.tex
\section{The \dataset Benchmark}
\label{sec:citeresolve}

Detecting an incorrect citation and establishing its source require different evidence. \dataset makes this distinction measurable by pairing real-world hallucinated citations with verified source-resolution targets.
We evaluate detection and resolution together, so that success requires identifying the error and correctly determining what source, if any, can be recovered.
% A citation with the wrong publication venue may still be traced to a real paper through its title and authors, whereas one with a real author's name but an invented title may have no identifiable source.

\subsection{Benchmark Construction}
\label{sec:benchmark-construction}

We build on the real-world test set of CiteAudit~\citep{shi2026citeaudit}, denoted by $\mathcal{C}=\{(c_i,l_i)\}_{i=1}^{N}$, where $c_i$ is a citation and $l_i\in\{\texttt{Real},\texttt{Fake}\}$ is its label.
\texttt{Real} denotes a citation that is valid as written, whereas \texttt{Fake} covers nonexistent works and substantive errors in comparison with real works.
We retain the full dataset and its released labels for evaluating hallucinated citation detection.

To evaluate source resolution, we further annotate CiteAudit's \texttt{Fake} subset.
We first determine whether a citation can be traced to a unique real work.
If so, we label it \texttt{Recoverable} and provide a verified corrected citation $c_i^\star$.
Otherwise, we label it \texttt{Unrecoverable} and set $c_i^\star=\texttt{None}$.
(Note that CiteAudit provides neither these recoverability labels nor the corrected citations.)
Let $\mathcal{I}\subseteq\{i:l_i=\texttt{Fake}\}$ index the cases retained after manual review and input filtering.
%as detailed in Appendix~\ref{app:citeresolve-construction}.
The resulting benchmark, $\widetilde{\mathcal{C}}=\{(c_i,c_i^\star)\}_{i\in\mathcal{I}}$, contains 423 citations: 252 \texttt{Recoverable} and 171 \texttt{Unrecoverable}.
Table~\ref{tab:citeresolve-stats} summarizes their source types and error patterns.

\begin{table}[t]
\centering
\small
\setlength{\tabcolsep}{3pt}
\begin{tabular}{@{}lrrr@{}}
\toprule
Cited as & Recoverable & Unrecoverable & Total \\
\midrule
Paper & 225 & 163 & 388 \\
Book & 3 & 1 & 4 \\
Web resource$^{\dagger}$ & 24 & 7 & 31 \\
\midrule
Total & 252 & 171 & 423 \\
\bottomrule
\end{tabular}

\vspace{8pt}
\begin{tabular}{@{}lr@{}}
\toprule
What a recoverable citation gets wrong & Citations \\
\midrule
Authors only & 185 (73\%) \\
Title only & 8 (3\%) \\
Title and authors & 33 (13\%) \\
Publication details & 15 (6\%) \\
Web page details & 11 (4\%) \\
\bottomrule
\end{tabular}
\caption{\dataset at a glance. $^{\dagger}$Web pages, datasets, software, and technical reports. Publication details refer to venue, year, volume, pages, or link. Webpage details involve the page title, the author or organization, and the address.
}
\label{tab:citeresolve-stats}
\vspace{-1em}
\end{table}

\subsection{Annotation Criteria}
\label{sec:resolution-targets}

Annotation separates evidence of an error from evidence of source identity.
Following reference-matching studies~\citep{visser2021large,han2026phantom}, we examine title similarity together with author and identifier evidence, allowing minor formatting differences.
The criteria vary with the type of source being cited.

\smallskip
\noindent \textbf{Papers.}
A near-exact title match is the main basis for recovery, allowing differences in capitalization, punctuation, and minor typographical errors.
When the title is very close in meaning but not a near-exact match, we additionally require an identifier in the citation, such as an arXiv ID or URL, to point to the candidate work.
An incorrect author list does not prevent recovery if the remaining evidence uniquely identifies that work.
Year differences alone do not determine recoverability, since online and print publication dates can differ~\citep{aria2026multiphase}.
Note that these rules concern whether the source can be recovered. Incorrect fields still need to be corrected in $c^\star$.

\smallskip
\noindent \textbf{Books and Web Resources.}
For books, we require agreement in both title and authors.
Minor publisher discrepancies do not by themselves rule out recovery.
For web resources, we follow the cited URL and compare the page title and any available author or organization information.
A page without an identifiable author is not rejected for that reason alone, consistent with reference conventions for works without author information\footnote{\url{https://apastyle.apa.org/style-grammar-guidelines/references/missing-information}}.
An explicit author mismatch, however, remains a discrepancy to resolve.

\smallskip
\noindent \textbf{\texttt{Unrecoverable} Citations.}
We assign $c^\star=\texttt{None}$ when no candidate satisfies these criteria or when several distinct works remain plausible.
Thus, \texttt{None} denotes the absence of a uniquely supported correction under our criteria, rather than a claim that no related work exists.

\subsection{Evaluation Protocol}
\label{sec:resolution-evaluation}

We run the evaluated algorithm on every citation $c_i$ in $\mathcal{C}$, providing \textit{only} the citation string.
It first predicts a label $\hat{l}_i$ and, when $\hat{l}_i=\texttt{Fake}$, determines whether the citation can be recovered.
If it judges the citation \texttt{Recoverable}, it returns a proposed corrected citation $\hat{c}_i^*$.
If it judges the citation \texttt{Unrecoverable}, it returns $\hat{c}_i^*=\texttt{None}$.

Detection is evaluated over all of $\mathcal{C}$ against $l_i$, with accuracy defined as
\begin{equation}
\frac{1}{|\mathcal{C}|}\sum_{i=1}^{|\mathcal{C}|}\mathbf{1}[\hat{l}_i=l_i].
\label{eq:detection-accuracy}
\end{equation}

% We also report precision, recall, and F1 with \texttt{Fake} as the positive class.

Resolution is evaluated over the citations indexed by $\mathcal{I}$.
A successful outcome requires both a correct \texttt{Fake} prediction and a correct recovery or abstention:
\begin{equation}
\frac{1}{|\mathcal{I}|}\sum_{i\in\mathcal{I}}
\mathbf{1}[\hat{l}_i=\texttt{Fake}]\,
\mathrm{Match}(\hat{c}_i^*,c_i^\star).
\label{eq:resolution-success}
\end{equation}
Here, $\mathrm{Match}(\hat{c}_i^*,c_i^\star)=1$ if the predicted correction is judged correct under the procedure in Section~\ref{sec:setup}, or if both values are \texttt{None}.
Otherwise, it is $0$.
Predicting \texttt{Real} always counts as a resolution failure on this subset, so missed detections are penalized in the end-to-end score.

%% file: 4_method.tex
\section{The \method Framework}
\label{sec:method}

\begin{figure*}[t]
\centering
\includegraphics[width=\textwidth]{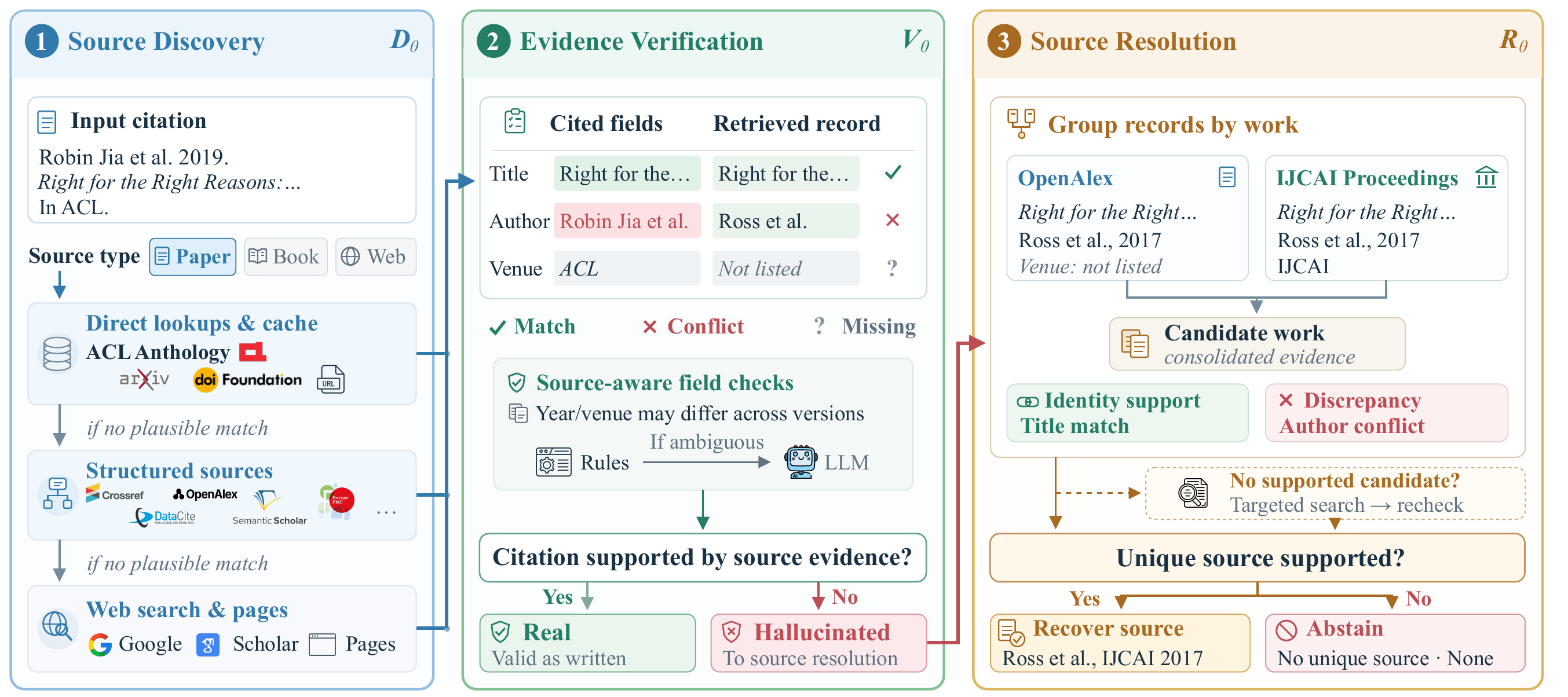}
\caption{
Overview of \method with a running example.
\method retrieves source-specific evidence, verifies citation fields against the retrieved records, and resolves the underlying source when one work is uniquely supported.
Otherwise, it abstains from source resolution.
}
\label{fig:citeanchor-overview}
\vspace{-1em}
\end{figure*}

\method implements \emph{source-aware grounding} by tailoring retrieval tools and bibliographic checks to each source type.
Its \emph{evidence decomposition} separates field comparisons, citation validity, and source identity, so that a matching field does not validate the entire citation and an erroneous field does not automatically rule out its source.
Inspired by the chain-of-evidence principle~\citep{meng2026scientistone}, our chain-of-source strategy connects three modules: \textbf{source discovery} $\mathsf{D}_{\theta}$, \textbf{evidence verification} $\mathsf{V}_{\theta}$, and \textbf{source resolution} $\mathsf{R}_{\theta}$.
They combine deterministic checks with LLM reasoning and share retrieved records, where $\theta$ denotes the common backbone used for their LLM calls. Figure~\ref{fig:citeanchor-overview} gives an overview of \method.

\subsection{Source Discovery}
\label{sec:source-discovery}

Given a citation $c$, the source discovery module $\mathsf{D}_{\theta}$ extracts its title, authors, identifiers, venue, and year when available.
It uses identifiers for direct lookups and title/author information to construct search queries.
It also identifies whether the citation refers to a paper, book, or web resource, which determines the relevant retrieval tools.
Starting with $\mathcal{E}_0=\varnothing$, $\mathsf{D}_{\theta}$ retrieves records $\mathcal{R}_t$ and accumulates evidence $\mathcal{E}_t$:
\begin{equation}
\mathcal{R}_t=\mathsf{D}_{\theta}(c,\mathcal{E}_{t-1}), \quad
\mathcal{E}_t=\mathcal{E}_{t-1}\cup\mathcal{R}_t.
\label{eq:evidence-collection}
\end{equation}
Retrieval first checks cached records and local resources, such as an offline copy of the ACL Anthology~\citep{bird-etal-2008-acl}.
It then follows available identifiers and queries structured bibliographic services, such as Crossref~\citep{hendricks2020crossref}.
Broader web search, such as Google Search (through SerpApi\footnote{\url{https://serpapi.com/search-api}}), is used only when these sources fail to provide a plausible matching record.
The concrete source configuration is given in Section~\ref{sec:setup}.

When incorrect fields obscure a match, $\mathsf{D}_{\theta}$ can revise the query, for example by searching by title after author information leads to conflicting results.

\subsection{Evidence Verification}
\label{sec:source-verification}

Given $c$ and the retrieved evidence $\mathcal{E}$, the evidence verification module $\mathsf{V}_{\theta}$ compares the cited fields with available records, distinguishing agreement, contradiction, and missing evidence.
If a retrieved record omits a field, it cannot confirm whether the corresponding value in the citation is correct or not.
Conversely, omitting an optional field from the citation is not itself an error.
The comparison also accounts for source type and legitimate year or venue differences between publication versions~\citep{visser2021large,shi2026citeaudit}.
Deterministic checks handle clear matches and contradictions, and the LLM assesses cases that these checks do not settle.
The resulting validity decision is
\begin{equation}
\hat{l}=\mathsf{V}_{\theta}(c,\mathcal{E}), \quad
\hat{l}\in\{\texttt{Real},\texttt{Fake}\}.
\end{equation}
For citations predicted as \texttt{Fake}, the evidence is passed to source resolution to either recover a uniquely supported source or conclude that no unique source can be established.

\subsection{Source Resolution}
\label{sec:source-resolution}

For a citation $c$ predicted to be hallucinated, the source resolution module $\mathsf{R}_{\theta}$ uses the collected evidence $\mathcal{E}$ to establish which real work, if any, remains supported.
It groups records of the same work and represents each group by one candidate citation.
The field comparisons distinguish evidence supporting a candidate's identity from discrepancies that require correction.
For example, a contradicted author list can be corrected when the title and other available clues still identify the work.
Let $\mathcal{U}(c,\mathcal{E})$ denote the candidates that the evidence supports as possible sources of $c$.
If this set is empty, the module makes one additional targeted web search through SerpApi.
The returned records $\mathcal{R}_{\mathrm{extra}}$ are added to the evidence set: $\mathcal{E}\leftarrow\mathcal{E}\cup\mathcal{R}_{\mathrm{extra}}$.
The module then checks the candidates against all available evidence and returns $\hat{c}^*=\mathsf{R}_{\theta}(c,\mathcal{E})$, with
\begin{equation}
\hat{c}^*=
\begin{cases}
x, & \mathcal{U}(c,\mathcal{E})=\{x\},\\
\texttt{None}, & \text{otherwise}.
\end{cases}
\label{eq:source-resolution}
\end{equation}
A unique supported candidate becomes the proposed corrected citation.
If no candidate is supported or several distinct works remain plausible, \texttt{None} withholds a correction while retaining the \texttt{Fake} judgment.

%% file: 5_experiment.tex
\section{Experiments}

\subsection{Experimental Setup}
\label{sec:setup}

\noindent \textbf{Datasets.}
We evaluate detection on CiteAudit and source resolution on \dataset, following the end-to-end protocol in Section~\ref{sec:resolution-evaluation}.

\smallskip
\noindent \textbf{Backbones.}
To demonstrate that \method supports different backbones, we instantiate it with the proprietary GPT-5.6 Luna~\citep{openai2026gpt56luna} and the open-weight OpenResearcher-30B-A3B~\citep{li2026openresearcher}.
The latter is post-trained on multi-step search and browsing trajectories, making it a suitable choice for agentic information seeking.
Including such an open-weight backbone also allows local model inference when submission confidentiality restricts the use of external LLM services.

\begin{table*}[t]
\centering
\small
\setlength{\tabcolsep}{4pt}
\begin{tabularx}{\textwidth}{@{}l*{8}{>{\centering\arraybackslash}X}@{}}
\toprule
& \multicolumn{4}{c}{Confusion matrix} & \multicolumn{4}{c}{Metrics} \\
\cmidrule(lr){2-5}\cmidrule(lr){6-9}
Method & TP & FN & FP & TN & Acc. & Prec. & Rec. & F1 \\
\midrule

\rowcolor{refbg}
\multicolumn{9}{c}{\textit{Published results for reference~\citep{shi2026citeaudit}}} \\
\midrule
GPTZero\reported~\cite{adam2026gptzerorobustdetectionllmgenerated}
& 338 & 129 & 1,358 & 1,531
& 0.557 & 0.199 & 0.724 & 0.313 \\
Gemini-3-Pro\reported~\cite{google2025gemini3pro}
& 351 & 116 & 412 & 2,477
& 0.843 & 0.460 & 0.752 & 0.571 \\
GPT-5.2\reported~\cite{openai2025gpt52}
& 366 & 101 & 1,379 & 1,510
& 0.559 & 0.210 & 0.784 & 0.331 \\
Claude-Sonnet-4.5\reported~\cite{anthropic2025sonnet45}
& 349 & 118 & 756 & 2,133
& 0.740 & 0.316 & 0.747 & 0.444 \\
CiteAudit\reported{}~\cite{shi2026citeaudit}
& 444 & 23 & 149 & 2,740
& 0.949 & 0.749 & 0.951 & 0.838 \\

\midrule

\rowcolor{backbonebg}
\multicolumn{9}{c}{\textit{Backbone: GPT-5.6 Luna~\citep{openai2026gpt56luna}}} \\
\midrule
Direct Prompting
& 120 & 347 & 54 & 2,835
& 0.881 & 0.690 & 0.257 & 0.374 \\
CiteAudit~\cite{shi2026citeaudit}
& 361 & 106 & 120 & 2,769
& 0.933 & 0.751 & 0.773 & 0.762 \\
\textbf{\method}
& 378 & 89 & 54 & 2,835
& \textbf{0.957} & \textbf{0.875} & \textbf{0.809} & \textbf{0.841} \\

\midrule

\rowcolor{backbonebg}
\multicolumn{9}{c}{\textit{Backbone: OpenResearcher-30B-A3B~\citep{li2026openresearcher}}} \\
\midrule
Direct Prompting
& 100 & 367 & 497 & 2,392
& 0.743 & 0.168 & 0.214 & 0.188 \\
CiteAudit~\cite{shi2026citeaudit}
& 222 & 245 & 379 & 2,510
& 0.814 & 0.369 & 0.475 & 0.416 \\
\textbf{\method}
& 384 & 83 & 128 & 2,761
& \textbf{0.937} & \textbf{0.750} & \textbf{0.822} & \textbf{0.784} \\

\bottomrule
\end{tabularx}
\caption{Detection on the CiteAudit real-world test set.
Published results (\reported) from \citet{shi2026citeaudit} are included for reference.
CiteAudit's published result uses Gemini-3-Flash~\citep{google2025gemini3flash}.
The lower blocks compare methods using the indicated backbone.
\textbf{Bold} marks the best metric within each backbone block.}
\label{tab:citeaudit-main}
\vspace{-1em}
\end{table*}

\smallskip
\noindent \textbf{Baselines.}
For detection, we include the results reported by \citet{shi2026citeaudit} for GPTZero~\citep{adam2026gptzerorobustdetectionllmgenerated}, Gemini-3-Pro~\citep{google2025gemini3pro}, GPT-5.2~\citep{openai2025gpt52}, Claude-Sonnet-4.5~\citep{anthropic2025sonnet45}, and CiteAudit~\citep{shi2026citeaudit} with the Gemini-3-Flash backbone~\citep{google2025gemini3flash}.
These published results serve as reference points and are not rerun.
With both backbones, we run \textbf{Direct Prompting}, which answers without retrieval, and \textbf{CiteAudit}~\citep{shi2026citeaudit}, which uses the released pipeline.
For the CiteAudit reruns, we replace its original backbone while retaining the released prompts.
For resolution, Direct Prompting predicts citation validity and returns a source or \texttt{None} for citations it flags.
\textbf{Web Search} uses the same prompt but permits web search through OpenAI's web-search tool for GPT-5.6 Luna~\citep{openai2026gpt56luna} or OpenResearcher's browser tools~\citep{li2026openresearcher}.
Implementations are provided in Appendix~\ref{app:baselines}.

\smallskip
\noindent \textbf{Configuration.}
\label{config.}
Retrieval first checks records cached within the current run and the offline ACL Anthology~\citep{bird-etal-2008-acl}, then follows identifiers and queries source-specific services.
Identifiers include DOIs, arXiv IDs, and cited URLs.
The services include Crossref~\citep{hendricks2020crossref}, OpenAlex~\citep{priem2022openalexfullyopenindexscholarly}, arXiv~\citep{ginsparg2011arxiv}, Semantic Scholar~\citep{lo2020s2orc}, Europe PMC~\citep{europe2015europe}, Google Books, and DataCite~\citep{datacite2026metadata}.
Google Scholar and Google Search through SerpApi are used when earlier sources provide no plausible match.
Deterministic checks handle clear cases, while unresolved comparisons go to the LLM judge.
Resolution reuses the collected evidence and adds at most one targeted search when no candidate is supported.
The same retrieval rules and verification thresholds apply to both backbones.
Appendices~\ref{app:impl-discovery}--\ref{app:impl-resolution} specify retrieval routes, decision rules, and inference settings.

\smallskip
\noindent \textbf{Evaluation Metrics.}
For detection, following \citet{shi2026citeaudit}, we report \textit{accuracy} (as defined in Section~\ref{sec:resolution-evaluation}), \textit{precision}, \textit{recall}, and \textit{F1}, treating \texttt{Fake} as positive.
For resolution, we report \emph{Identified (Recoverable)}: the fraction of the 252 recoverable citations predicted \texttt{Fake} with a proposed source.
\emph{Recovered} measures conditional recovery accuracy: among these proposals, the fraction that matches the gold source.
\emph{Identified (Unrecoverable)} is the fraction of the 171 unrecoverable citations predicted \texttt{Fake} with \texttt{None}.
\emph{Task success} (as defined in Section~\ref{sec:resolution-evaluation}) measures correct detection followed by correct recovery or abstention across all 423 citations.
Recovered sources are compared with gold records by title, authors, venue, and year, allowing formatting differences and legitimate publication variants.
Deterministic rules resolve clear matches, while GPT-6 Sol~\citep{openai2026gpt6sol} judges ambiguous fields without access to the original citation or system identity.
Appendix~\ref{app:eval} details the matching procedure.

\subsection{Citation Detection on the CiteAudit Real-World Test Set}

\noindent \textbf{Overall Results.}
Table~\ref{tab:citeaudit-main} compares all methods on the same 3,356 citations.
The published results provide context, while the lower blocks compare methods using the same backbone.
With GPT-5.6 Luna, \method improves F1 over CiteAudit from 0.762 to 0.841, a gain of 7.9 points.
Both precision (0.875 vs.\ 0.751) and recall (0.809 vs.\ 0.773) improve.
The gain is larger with OpenResearcher-30B-A3B, where F1 rises from 0.416 to 0.784.
Here, \method reduces false positives from 379 to 128 and missed hallucinated citations from 245 to 83.
These improvements show that source-aware grounding benefits both proprietary and open-weight backbones.

\begin{table*}[t]
\centering
\small
\setlength{\tabcolsep}{4pt}
\begin{tabularx}{\textwidth}{@{}l*{4}{>{\centering\arraybackslash}X}@{}}
\toprule
& \multicolumn{2}{c}{Recoverable (252)} & \multicolumn{1}{c}{Unrecoverable (171)} & \multicolumn{1}{c}{Overall} \\
\cmidrule(lr){2-3}\cmidrule(lr){4-4}\cmidrule(lr){5-5}
Method & Identified~$\uparrow$ & Recovered~$\uparrow$ & Identified~$\uparrow$ & Task success~$\uparrow$ \\
\midrule
\rowcolor{backbonebg}
\multicolumn{5}{c}{\textit{Backbone: GPT-5.6 Luna~\citep{openai2026gpt56luna}}} \\
\midrule
Direct Prompting & 23.0 & 36.2 & 19.3 & 12.8 \\
Web Search & \textbf{85.3} & 90.2 & 31.0 & 58.4 \\
\textbf{\method} & 65.9 & \textbf{95.2} & \textbf{92.4} & \textbf{74.7} \\
\midrule
\rowcolor{backbonebg}
\multicolumn{5}{c}{\textit{Backbone: OpenResearcher-30B-A3B~\citep{li2026openresearcher}}} \\
\midrule
Direct Prompting & 25.8 & 12.3 & 3.5 & 3.3 \\
Web Search & 61.9 & 10.9 & 3.5 & 5.4 \\
\textbf{\method} & \textbf{68.7} & \textbf{87.9} & \textbf{86.0} & \textbf{70.7} \\
\bottomrule
\end{tabularx}
\caption{Source resolution on \dataset under the end-to-end protocol in Section~\ref{sec:resolution-evaluation}.
All values are percentages.
Metrics are detailed in Section~\ref{sec:setup}.
\textbf{Bold} marks the best value within each backbone block.}
\label{tab:citeresolve-main}
\vspace{-0.5em}
\end{table*}

\begin{figure}[t]
\centering
\includegraphics[width=\columnwidth]{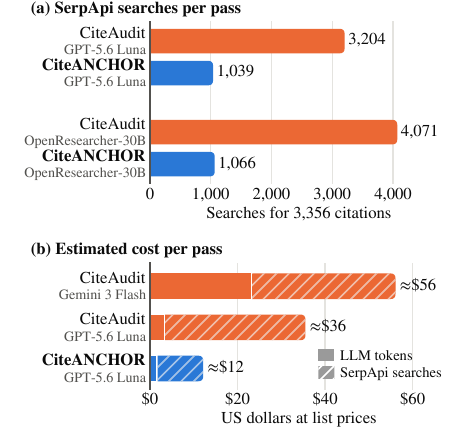}
\caption{
Commercial search use and estimated API cost over the 3,356 CiteAudit citations.
(a) Number of SerpApi searches, including additional searches during source resolution.
(b) Estimated combined cost of LLM calls and SerpApi searches using public API pricing.
Hatching indicates search costs.
}
\label{fig:serpapi}\label{fig:cost}
\vspace{-1em}
\end{figure}

\smallskip
\noindent \textbf{Efficiency and Cost.}
Source-aware retrieval also reduces reliance on commercial search.
Over the 3,356 citations, SerpApi calls decrease from 3,204 to 1,039 with GPT-5.6 Luna and from 4,071 to 1,066 with OpenResearcher-30B-A3B, reductions of 67.6\% and 73.8\%, respectively (Figure~\ref{fig:cost}).
With GPT-5.6 Luna, the estimated combined cost of LLM calls and commercial searches falls from approximately \$36 to \$12 per run.

\subsection{Source Resolution on \dataset}

\noindent \textbf{Overall Results.}
Table~\ref{tab:citeresolve-main} evaluates both recovering a uniquely supported source and withholding a correction when no such source can be established.
With GPT-5.6 Luna, Web Search proposes sources for more recoverable citations than \method (85.3\% vs.\ 65.9\%).
However, \method's proposed sources are more accurate (95.2\% vs.\ 90.2\%), and it correctly identifies far more unrecoverable citations (92.4\% vs.\ 31.0\%).
This combination yields 74.7\% overall task success, exceeding Web Search by 16.3 percentage points despite lower recovery coverage.

The advantage is larger with OpenResearcher-30B-A3B.
\method performs best on all four metrics and reaches 70.7\% task success, compared with 5.4\% for Web Search and 3.3\% for Direct Prompting.
These results show that access to web search alone does not ensure reliable source resolution with this backbone.

%上标一个颜色/#号，上面的表, 强调是抄来的，不止一处可以强调；可以弱化；颜色加上，加粗在下面，caption下面说一下是三个之间比较即可；price

%price：在下面专门写基模型的钱的对比，主表就是performance

%把reference可以加上，
%不知道的就不比了，serp api可以画一个图啥的，两个东西比就行了；

%ours也要标注好名字；

%citeResolve, 
%太稀疏：加evaluation metric, 

%等到dealine前10天，或者说

%尝试一个开源模型，qwen
%GPT-OSS-20B Nemotron-30B-A3B Gemma-35B-A4B
%开源试试看，totally broken, 能有一个还行的结果，
% OpenResearcher 30B 试试看

%如果说，至少两个backbone, claude+开源 backbone, 
% claude 

% Abaltion
% Analysis

% story, 
% contribution: benchmark 肯定是有的, 一些方法降低serp api, 

% idea: multi-agent, RSI, Memory, chain of source, 想一想, 归归类, web/database, chain里面有agent, agentic, 
% orchetractor

% performance: evidence, 

% title, 不卖benchmark, 

% to do:
% 开源model, 能不能放进去, performance到0.6就可以了,
% chain of source, 往上贴, 
% abaltion根据story

%图

%表有点多，图有点少，Benchmark的可以看看整一个图啥的

%api的图比较占空间
% cost的实际开销
% 累计的总开销的图画一个，标注预估

% figure 2 保留， figure 1换掉，换成一个合成的表

%% file: 6_analysis.tex
\section{Analysis}

\subsection{Ablation Studies}
\label{sec:cos-ablation}

We examine three design choices in \method: evidence decomposition, complementary retrieval sources, and source-type adaptation.
Table~\ref{tab:ablation} reports detection F1 on CiteAudit and resolution success on \dataset with GPT-5.6 Luna.

\smallskip
\noindent \textbf{Evidence Decomposition.}
We first compare the full verification and resolution procedure with a single joint judgment over the same retrieved records.
The \emph{w/o Chain} variant asks the backbone to decide citation validity and source identity in one call, using the prompt in Appendix~\ref{app:prompts-baselines}.
Detection F1 falls from 0.841 to 0.643, and resolution success from 74.7\% to 59.3\%.
The comparison evaluates the decision procedure as a whole, including its field checks and instructions.
It supports the benefit of explicitly checking citation fields and separating validity from source identity, even when the retrieved evidence is held fixed.

\begin{table}[!htb]
\centering
\small
\setlength{\tabcolsep}{3.5pt}
\begin{tabular}{lcc}
\toprule
Variant & CiteAudit F1 & CiteResolve Success \\
\midrule
w/o Chain & 0.643 & 59.3 \\
\midrule
w/o Search Engines & 0.734 & 72.6 \\
w/o Structured APIs & 0.773 & 53.2 \\
\midrule
w/o Type Adaptation & 0.761 & 66.9 \\
\midrule
\textbf{\method} & \textbf{0.841} & \textbf{74.7} \\
\bottomrule
\end{tabular}
\caption{Ablations with GPT-5.6 Luna.
Resolution success is reported as a percentage.
\emph{w/o Chain} replaces staged verification and resolution with one LLM judgment over the same records.
\emph{w/o Search Engines} disables Google and Google Scholar searches.
\emph{w/o Structured APIs} removes bibliographic APIs while retaining search engines and cited webpages.
\emph{w/o Type Adaptation} uses generic retrieval and verification rules for all source types.}
\label{tab:ablation}
\vspace{-1em}
\end{table}

\smallskip
\noindent \textbf{Retrieval Sources.}
Removing search engines lowers detection F1 from 0.841 to 0.734, while resolution success decreases more modestly, from 74.7\% to 72.6\%.
Removing structured APIs also hurts both tasks, but the larger loss is in resolution: success falls to 53.2\%, while detection F1 remains at 0.773.
The different patterns suggest complementary roles for the two source families.
Broad web search helps detect citation errors, while structured bibliographic records are particularly valuable for establishing which source can be recovered.

\smallskip
\noindent \textbf{Source-Type Adaptation.}
Finally, replacing type-specific retrieval and verification with a generic procedure reduces detection F1 to 0.761 and resolution success to 66.9\%.
Papers, books, and web resources differ in where they can be found and what evidence identifies them. These decreases support adapting both evidence collection and interpretation to the source type.

\subsection{Case Studies}
\label{sec:case-study}
Figure~\ref{fig:case-study} illustrates why verification must consider both what a retrieved record says and whether it describes the cited work itself.
In Case (a), the title identifies a real paper, but seven of the ten cited authors are absent from its complete 13-author list.
All three baselines accept the citation.
CiteAudit attributes the discrepancy to truncated or differing records, while Web Search and Direct Prompting incorrectly state that the author list matches.
\method detects the author contradiction while using the matching title to recover the source.
The same evidence thus supports a \texttt{Fake} label and a recoverable source.

In Case (b), the citation appears on the web, but the retrieved bibliographic records do not establish a matching source.
Its arXiv identifier resolves to a different paper.
All three baselines again accept the citation, and CiteAudit's explanation specifically cites a webpage reproducing its title and authors.
\method rejects the mismatched identifier and returns \texttt{Fake} with \texttt{None}.
This case illustrates why a webpage that repeats a reference is insufficient evidence that the cited work exists.

\input{figures/fig_cases_baselines}

% 明天中午之前决定是否跑claude api, 帮跑一下, 
% 换个背景颜色，背景换成灰色的，下面换成全是蓝色的
% 表格里面把ref也加上
% 先写 benchmark, bench写在method前面

% benchmark的一些基本的统计结果
% 完整的表可放到appendix

% related work 0.75 page, 一列多 
% 定量:
% ablation: 需要全集
% 1. evidence buffer, chain 
% 2. 等claude backbone出来再说
% 3. search data质量, source
% 定性: case study, 找一下
% appendix 堆一些细节, 有争议的不要写, 

% appendx prompt 放一些
% more case study

% 明天中午, 跑完claude, 
% 下周二三 把表1/2 加上三行
% case baseline没fix 我们fix

%% file: figures/fig_cases_baselines.tex
% Figure: two CiteResolve citations on which every baseline fails (compact version, 2026-10-11).
% (a) CiteResolve id 150 / CiteAudit row 2017, (b) id 137 / row 1911. Baseline answers from
% baselines/results/realworld_all_citeaudit.json, baselines/results/citeresolve/citeresolve_websearch1_luna_v2.jsonl,
% baselines/results/realworld3356_direct_luna.json; our records/rules from results1/realworld3356_citeanchor_luna_v9.json
% and ablation/source/scholar_venue/replay_FULL_A_SV.json. Long version: archive/fig_cases_baselines.2026-10-11-long.tex.
\begin{figure}[t]
\centering
\providecolor{oursbg}{HTML}{E8F1FB}
\tikzset{
  every node/.style={font=\scriptsize, align=left},
  cit/.style={draw=black!35, rounded corners=2pt, inner sep=3pt, text width=7.3cm},
  chip/.style={draw=red!50, fill=red!4, rounded corners=2pt, inner sep=2.5pt, text width=2.25cm,
    execute at begin node={\hyphenpenalty=10000\exhyphenpenalty=10000}},
  chipn/.style={chip, text width=2.0cm},    % middle chip (short quotes)
  chipw/.style={chip, text width=2.55cm},   % third chip, wide enough for its longer name
  ours/.style={draw=blue!35, fill=oursbg, rounded corners=2pt, inner sep=3pt, text width=7.3cm},
  final/.style={draw=green!40!black, fill=green!6, rounded corners=2pt, inner sep=3pt, text width=7.3cm},
  hd/.style={font=\scriptsize\bfseries, inner sep=1pt},
}
\newcommand{\ok}{\textcolor{green!45!black}{\checkmark}}
\newcommand{\no}{\textcolor{red!70!black}{$\times$}}
\newcommand{\bad}[1]{\textcolor{red!70!black}{#1}}
\newcommand{\sys}[3][black!60]{{\fontsize{7}{8}\selectfont{\color{#1}#2}~\bfseries #3}}   % system names 7pt so that "Direct Prompting" fits
\newcommand{\sysCA}{\sys[teal!70!black]{\faClipboardCheck}{CiteAudit}}
\newcommand{\sysWS}{\sys[blue!60!black]{\faSearch}{Web Search}}
\newcommand{\sysEE}{\sys[violet!80!black]{\faRobot}{Direct Prompting}}
\newcommand{\sysOurs}{\sys[blue!70!black]{\faAnchor}{\method}}
\begin{tikzpicture}[baseline=(t.base)]
\node[hd] (t) {(a) A real work cited with invented authors};
\node[cit, below=3pt of t] (c) {Raphael J L Townshend, Stephan Eismann, \bad{Andrew M Watkins, Rishab Rangan, Maria
  Karelina, Rajesh Das, John Jumper, Gerard J Kleywegt, Ashesh Krishnapriyan}, and Ron O Dror. Atom3d: Tasks on
  molecules in three dimensions. In NeurIPS, 2021.};
\node[chip, below=4pt of c.south west, anchor=north west] (b1) {\sysCA{}\,\no\\
  Real: ``attributable to differing/\allowbreak truncated metadata''};
\node[chipn, right=3pt of b1.north east, anchor=north west] (b2) {\sysWS{}\,\no\\ Real: ``author list
  correspond to the real \ldots\ work''};
\node[chipw, right=3pt of b2.north east, anchor=north west] (b3) {\sysEE{}\,\no\\ Real: ``a real ATOM3D paper with the cited title and author list''};
\node[fit=(b1)(b2)(b3), inner sep=0pt] (row) {};
\node[ours, below=4pt of row.south west, anchor=north west] (o) {\sysOurs: title matches ATOM3D
  (arXiv 2020); 7 of the 10 cited authors are absent from its 13-author list. Author field contradicted; the title
  identifies the work.};
\node[final, below=4pt of o] (f) {\textbf{Hallucinated} \ok\quad \textbf{Recoverable} \ok\quad Source: Townshend,
  V\"ogele, \ldots, Dror. ATOM3D: Tasks On Molecules in Three Dimensions. 2020.};
\end{tikzpicture}
\\[6pt]
\begin{tikzpicture}[baseline=(t.base)]
\node[hd] (t) {(b) A fabricated citation that search engines find};
\node[cit, below=3pt of t] (c) {\bad{Chen Zhu et al. A survey on efficient deployment of large language models.} arXiv
  preprint arXiv:2307.03744, 2023.};
\node[chip, below=4pt of c.south west, anchor=north west] (b1) {\sysCA{}\,\no\\
  Real: a page ``reproduces the exact title and authors''};
\node[chipn, right=3pt of b1.north east, anchor=north west] (b2) {\sysWS{}\,\no\\ Real: ``a real
  paper \ldots\ with Chen Zhu as an author''};
\node[chipw, right=3pt of b2.north east, anchor=north west] (b3) {\sysEE{}\,\no\\ Real: ``a real 2023 arXiv work \ldots\ has Chen~Zhu as an author''};
\node[fit=(b1)(b2)(b3), inner sep=0pt] (row) {};
\node[ours, below=4pt of row.south west, anchor=north west] (o) {\sysOurs: arXiv:2307.03744 resolves to
  \emph{Comparing Traditional and LLM-based Search \ldots} (Spatharioti et al.): identifier ruled out. No bibliographic
  record carries the title, and pages that merely contain the citation do not establish the existence of a source.};
\node[final, below=4pt of o] (f) {\textbf{Hallucinated} \ok\quad \textbf{Unrecoverable} \ok\quad Source: None.};
\end{tikzpicture}
\caption{Two \dataset cases where CiteAudit, Web Search, and Direct Prompting fail with GPT-5.6 Luna.
Red text marks incorrect citation fields, and red boxes show excerpts from the baselines' explanations.
(a) A real title is paired with an incorrect author list.
\method detects the contradiction and recovers the source.
(b) An unrecoverable citation is repeated on the web without a supporting bibliographic record.
\method rejects the mismatched identifier and returns \texttt{None}.}
\label{fig:case-study}
\vspace{-1em}
\end{figure}
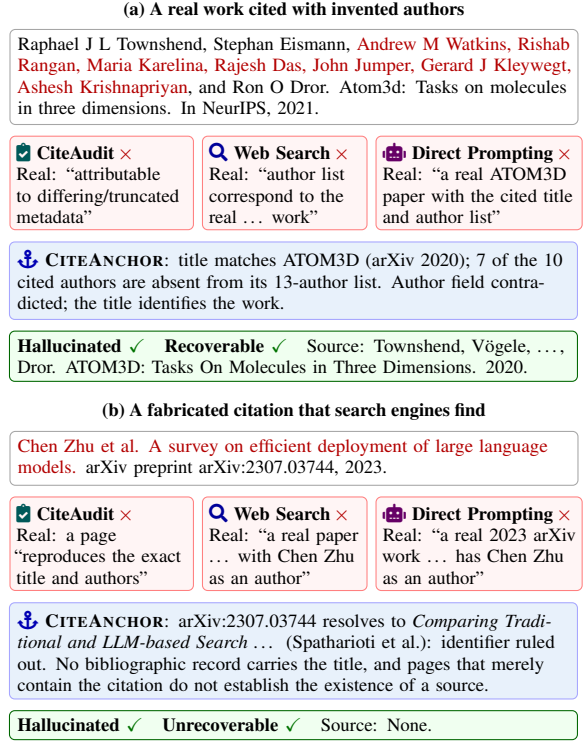

%% file: 7_conclusion.tex
\section{Conclusion}
We introduce \dataset, a benchmark for source resolution in hallucinated citations, and \method, a source-aware grounding framework that separates citation validity from source identity.
Across proprietary and open-weight backbones, \method improves hallucinated citation detection and end-to-end source resolution while reducing commercial search use.
Our results suggest that reliable citation verification should go beyond detecting errors to recover the underlying source when uniquely supported, and abstain from correction otherwise.
This helps human reviewers inspect detected errors and assess proposed corrections using evidence from retrieved records.

\section*{Limitations}
First, \dataset is dominated by citations to papers, with fewer books and web resources.
The reported results therefore provide stronger evidence for paper citations than for other source types.
Broader coverage across disciplines and languages would also help assess how well the framework transfers to other settings.
Second, our task concerns bibliographic validity and source identity.
The system receives only a citation entry and does not assess whether the cited work supports the surrounding claim.
A correct reference can still be used in an unsupported context, so bibliographic verification and claim-level attribution should be viewed as complementary tasks.
Finally, our evaluation reflects the records and services available at the time of the experiments.
Bibliographic databases and web pages change, and the same citation may become easier or harder to resolve as records are added or revised.
Accordingly, \texttt{Unrecoverable} means that no unique source is supported by the available evidence under our criteria, rather than that no related work exists.

\section*{Ethical Considerations}
\method is intended to help reviewers detect and examine problematic references.
It is not intended to let authors bypass reference checking or rely on automatic corrections without verification.
Authors remain responsible for the accuracy of every reference they include.
That said, a \texttt{Fake} prediction by \method concerns the citation itself and does not establish LLM use, deliberate fabrication, or author misconduct~\citep{sakai2026hallucitation}.

Our experiments use citations from the existing CiteAudit benchmark and public bibliographic sources.
Deployment on confidential submissions requires model and retrieval services that meet the venue's confidentiality requirements.
A locally hosted backbone (e.g., OpenResearcher-30B-A3B~\citep{li2026openresearcher} used in our experiments) can keep model inference local, but external searches still transmit queries to service providers.
Only the information needed for reference verification should be shared, subject to the applicable confidentiality rules and service terms. We used AI assistants for coding and for writing assistance.

%% file: 8_appendix.tex
% \clearpage
\appendix

\addtocontents{toc}{\protect\setcounter{tocdepth}{2}}
\renewcommand{\contentsname}{Appendix Contents}
\makeatletter
\renewcommand*\l@section[2]{\ifnum\c@tocdepth>\z@\addpenalty\@secpenalty\addvspace{1.15em \@plus\p@}%
  \setlength\@tempdima{1.5em}\begingroup\parindent\z@\rightskip\@pnumwidth\parfillskip-\@pnumwidth
  \leavevmode\bfseries\advance\leftskip\@tempdima\hskip-\leftskip#1\nobreak\hfil\nobreak\hb@xt@\@pnumwidth{\hss#2}\par\endgroup\fi}
\makeatother
\begingroup\linespread{1.1}\selectfont
\tableofcontents
\endgroup

\section{\dataset Details}
\label{app:citeresolve}

\subsection{Construction}
\label{app:citeresolve-construction}

\noindent \textbf{Starting Point.}
\dataset is constructed from citations labeled \texttt{Fake} in the CiteAudit real-world test set.
We retain the citation string from the original paper and manually annotate whether its source can be uniquely identified.
After annotation and quality control, the benchmark contains 423 citations.

Each retained citation is labeled \texttt{Recoverable} when the available evidence uniquely identifies a source, or \texttt{Unrecoverable} when no unique source can be established.

\smallskip
\noindent \textbf{Annotation Workflow.}
For each citation, the annotation tool first displayed the top three
records returned by Crossref for the full reference string and by OpenAlex
and Semantic Scholar for the parsed title. The returned records were pooled
without reranking. If none matched, the annotator could broaden the query,
resolve a DOI, arXiv ID, or URL, search by free text, or enter a source found
elsewhere. \method was not used during annotation. Annotation proceeded in two passes: one annotator labeled every citation that CiteAudit marks
as \texttt{Fake}, and a second annotator independently re-annotated all of them under the same
criteria; disagreements were resolved by discussion (see Quality Control below).

\smallskip
\noindent \textbf{Source-Identity Criterion.}
The annotation decision is based on whether the evidence identifies a
unique underlying source.
For papers, title evidence carries the most weight; an author list alone is
not sufficient to establish the intended work. Identifiers such as DOI,
arXiv ID, and URL are first resolved and then checked against the remaining
citation fields.

A citation is labeled \texttt{Recoverable} only when the available evidence points consistently to one source.
If no compatible source can be found or multiple sources remain plausible, it is labeled \texttt{Unrecoverable}.
For example, a partially corrupted title may match several real works while the author list points to another source.
Without a unique match, the citation remains \texttt{Unrecoverable}.

Different versions of the same underlying work, such as an arXiv preprint
and its later conference publication, are treated as the same source when
their identity is clear. Minor bibliographic differences, including year
differences across versions, a trailing ``et al.'' or ``others'' in the
author list, or fields that are missing on either side, do not by themselves
make a citation unrecoverable.

% \smallskip
% \noindent \textbf{Search Stopping Criterion.}
% TODO: Specify what evidence must be checked before assigning an
% unrecoverable label, and when manual search is stopped.
% Suggested content:
% - which retrieval channels are checked at minimum;
% - whether identifiers are always resolved when present;
% - when further search is considered unlikely to produce a new plausible source.

\smallskip
\noindent \textbf{Gold Sources.}
For each \texttt{Recoverable} citation, we record the verified source's title, authors, venue, year, and available identifier.
This record serves as the gold target for evaluation.

For web resources without standard bibliographic metadata, the gold source is the resolved webpage, dataset page, or repository rather than a complete
bibliographic record. We therefore leave unavailable fields empty and record the corrected source directly.

\smallskip
\noindent \textbf{Quality Control.}
After the initial annotation, all citations were checked again for extraction
and metadata errors. A second annotator independently re-annotated the
citations using the same source-identity criteria, without access to the first-pass labels,
and consulted the cited papers or authoritative web sources for uncertain cases.
On the 423 citations in the benchmark, the second annotator assigned a different
recoverability label to 8 citations and flagged 31 others as ambiguous under the
criteria; the two passes agree on 415 labels (Cohen's $\kappa = 0.96$). After
discussion, 6 labels were changed, the other 2 were kept, and the criteria were
clarified for the ambiguous cases.

\subsection{Data Analysis}
\label{app:citeresolve-analysis}

\noindent \textbf{Composition.}
Table~\ref{tab:citeresolve-stats} summarizes the 423 citations. Most citations refer to
papers (388), followed by web resources (31) and books (4). The web resources
include 17 webpages, 6 datasets, 6 reports, and 2 software
repositories. 

\begin{figure*}[t]
\centering
\includegraphics[width=\textwidth]{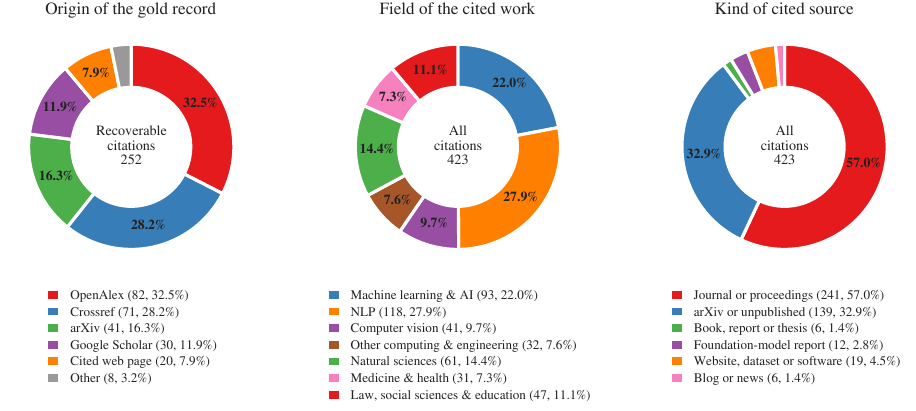}
\caption{
Additional statistics of \dataset.
Left: origin of the gold record for the 252 \texttt{Recoverable} citations.
``Cited web page'' denotes cases where the cited webpage itself
serves as the gold source; ``Other'' groups the remaining record sources.
Middle: research areas of all 423 citations.
Right: type of source cited by all 423 citations.
}
\label{fig:citeresolve-donuts}
\end{figure*}

Figure~\ref{fig:citeresolve-donuts} provides a broader view of the benchmark.
For the 252 \texttt{Recoverable} citations, most gold records come from
OpenAlex, Crossref, or arXiv, with Google Scholar and cited webpages covering
most of the remaining cases. The benchmark spans a range of research areas.
Computing accounts for the largest share, while the remaining citations cover
the natural sciences, medicine and health, and law, social sciences, and
education. Most cited sources are journal or conference papers or arXiv and
other unpublished works, with additional coverage of books, reports, websites,
datasets, software, blogs, and news articles.

\smallskip
\noindent \textbf{Identifiers.}
Table~\ref{tab:citeresolve-identifiers} shows that identifiers are common in
both classes. However, a valid identifier does not necessarily identify the
cited source. Among \texttt{Unrecoverable} citations, 35 of 39 arXiv IDs are
valid, yet none of the resolved papers matches the cited title. Identifiers
must therefore be resolved and checked against the remaining citation evidence.

\begin{table}[t]
\centering
\small
\begin{tabular}{lcc}
\toprule
 & \texttt{Recoverable} & \texttt{Unrecoverable} \\
\midrule
Contains identifier
    & 124 (49\%)
    & 59 (35\%) \\
Valid arXiv ID
    & 73 of 84 (87\%)
    & 35 of 39 (90\%) \\
Title match
    & 72 of 73 (99\%)
    & 0 of 35 (0\%) \\
\bottomrule
\end{tabular}
\caption{Identifier evidence in \dataset. The first row reports the share of
citations containing an arXiv ID, DOI, or URL. The next two rows focus on
citations containing an arXiv ID.}
\label{tab:citeresolve-identifiers}
\end{table}

\section{Implementation Details}
\label{app:impl}

This appendix specifies how the modules of Section~\ref{sec:method} are instantiated in the reported runs.
Appendix~\ref{app:impl-setup} gives the setup shared by all systems,
Appendices~\ref{app:impl-discovery}--\ref{app:impl-resolution} follow source discovery, evidence
verification, and source resolution, and Appendix~\ref{app:baselines} describes the baselines. Statistics
refer to the GPT-5.6 Luna run over the 3,356 CiteAudit citations.

\subsection{Common Setup}
\label{app:impl-setup}

\noindent \textbf{Backbones and Serving.}
Within each backbone comparison block of Tables~\ref{tab:citeaudit-main} and~\ref{tab:citeresolve-main}, all compared systems use the same backbone.
GPT-5.6 Luna is called through the OpenAI API with its default
sampling and reasoning settings and at most 4,096 output tokens per call, except for the CiteAudit rerun, which uses 8,192 (Appendix~\ref{app:baselines}). We serve OpenResearcher-30B-A3B locally with vLLM 0.13, as specified in the authors' implementation, using BF16 on one NVIDIA H20 per server, with a 32k context window and prefix caching. For \method, Direct Prompting, and the CiteAudit rerun, OpenResearcher's native browser tools are disabled.
Retrieval in \method and CiteAudit is handled by their respective pipelines.
OpenResearcher's reasoning budget is capped at 2,048 tokens, and responses are limited to 4,096 tokens and constrained to the required JSON schema.
\method uses the same rules and thresholds for both backbones.

\subsection{Source Discovery}
\label{app:impl-discovery}

\noindent \textbf{Fields and source type.}
$\mathsf{D}_{\theta}$ extracts the title, authors, identifier, venue, and year of a citation with one backbone call.
It then assigns a source type, which determines the retrieval route and the evidence used for verification.
We first apply deterministic bibliographic cues in a fixed order (Table~\ref{tab:type-rules}), covering
papers, books, reports, web pages, datasets, and software repositories. The LLM backbone is used when
these cues do not determine a type. Citations whose source type remains unresolved follow the paper route.

% This procedure classifies 3,285 of the 3,356 citations by rule, with
% the remaining 71 classified by the backbone.

\begin{table}[t]
\centering
\small
\setlength{\tabcolsep}{4pt}
\begin{tabularx}{\columnwidth}{@{}Xl@{}}
\toprule
Bibliographic cue & Type \\
\midrule
Academic identifier, preprint host, venue, volume/pages, or author--venue pattern
    & paper \\
ISBN, publisher, edition, handbook, encyclopedia, or chapter information
    & book \\
Technical/working paper, thesis, standard, or institutional report
    & report \\
Repository host such as GitHub, GitLab, Bitbucket, or Hugging Face model
    & repository \\
Dataset host such as Kaggle, Zenodo, figshare, Dryad, or OpenML
    & dataset \\
General webpage, blog, documentation, news, or accessed-date pattern
    & web page \\
% \midrule
% No cue applies
%     & backbone \\
 \bottomrule
\end{tabularx}
\caption{Deterministic cues used for source-type routing. Cues are applied in a fixed priority order;
the LLM backbone classifies only citations not resolved by these rules.}
\label{tab:type-rules}
\end{table}

\smallskip
\noindent \textbf{Type-Specific Retrieval.}
The source type determines the retrieval route and the sources consulted
(Table~\ref{tab:type-routes}). Papers are searched through bibliographic
services, books additionally through Google Books, and web resources first
at their cited address. Within each route, retrieval proceeds from direct
and structured sources to broader web search. The system first checks
previously retrieved records and an offline copy of the ACL Anthology,
then identifier lookups and structured bibliographic services, and uses
Google Scholar or Google Search only when the earlier sources do not yield a
plausible candidate. Retrieval broadens only when the current evidence
does not provide a usable candidate. Overall, 2,913 of the 3,356 citations
(87\%) required no commercial search.

% \begin{table*}[t]
% \centering
% \small
% \begin{tabularx}{\textwidth}{@{}>{\raggedright\arraybackslash}p{2.5cm}>{\raggedright\arraybackslash}X>{\raggedright\arraybackslash}p{2.6cm}>{\raggedright\arraybackslash}X@{}}
% \toprule
% Type & Searched first & Then & Accepted as a record of the cited work \\
% \midrule
% Paper & Identifier lookup; Crossref, OpenAlex, Semantic Scholar; arXiv when the citation points to a preprint; Europe PMC when it has a biomedical venue or term & Google Scholar, then Google & A bibliographic record with an author list. Not a web page, a Google hit, or a Scholar entry built only from other papers' reference lists. \\
% Book & Identifier lookup; the same services plus Google Books & Google Scholar, then Google & Any record, pages included; another edition counts as the same work \\
% Web resource (report, web page, dataset, repository) & The cited address, or an archived copy if it is gone; identifier lookup & Google, then Google Scholar & The page at the cited address, or any record at that address, if its title resembles the cited one \\
% Unknown & As for a paper & Google Scholar, then Google & Any record, pages included \\
% \bottomrule
% \end{tabularx}
% \caption{Source-aware retrieval and acceptance. The source type fixes where source discovery looks and
% which records evidence verification accepts as the cited work. Every route starts from the records
% retrieved earlier in the same run and the offline copy of the ACL Anthology.}
% \label{tab:type-routes}
% \end{table*}

\begin{table*}[t]
\centering
\small
\begin{tabularx}{\textwidth}{
@{}>{\raggedright\arraybackslash}p{2.2cm}
>{\raggedright\arraybackslash}X
>{\raggedright\arraybackslash}p{3.0cm}
>{\raggedright\arraybackslash}X@{}}
\toprule
Source type & Primary retrieval & Fallback retrieval & Evidence passed to verification \\
\midrule

Paper &
Identifier lookup and structured scholarly sources, including Crossref,
OpenAlex, Semantic Scholar, arXiv for preprints, and Europe PMC for
biomedical citations &
Google Scholar, then Google &
Bibliographic records with author metadata; generic web hits and
reference-list-only Scholar entries are excluded \\

Book &
Identifier lookup and scholarly sources, augmented with Google Books &
Google Scholar, then Google &
Book records whose title and author evidence are sufficient for
identity comparison; differences across editions are allowed \\

Web resource
% (report, web page, dataset, repository)

&
The cited address, or an archived copy when unavailable, together with
identifier lookup &
Google, then Google Scholar &
The cited page or records tied to that address whose title provides
evidence about the cited resource \\

% Unknown &
% The same structured sources used for papers &
% Google Scholar, then Google &
% Records satisfying the generic identity checks used in evidence
% verification \\

\bottomrule
\end{tabularx}

\caption{Source-aware retrieval routes. Source type determines the
primary evidence sources and the order in which retrieval is expanded.
All routes first consult previously retrieved records and the offline
ACL Anthology. Retrieved records are passed to evidence verification,
which determines whether they support the cited work.}
\label{tab:type-routes}
\end{table*}

\smallskip
\noindent \textbf{Query Revision.}
An incorrect citation field can prevent the intended source from appearing in the initial search.
When no plausible record is found, all retrieved candidates are ruled out as different works, or the
remaining evidence is too weak to verify the citation, $\mathsf{D}_{\theta}$ revises the query by dropping
or simplifying unreliable fields. For example, it may search by title alone or use a shortened title.
The revised search stops once a plausible record is found. In the GPT-5.6 Luna run, query revision was
triggered for 350 citations.

\subsection{Evidence Verification}
\label{app:impl-verification}

$\mathsf{V}_{\theta}$ compares a citation with the retrieved records field by field, combines the
resulting evidence into a validity decision, and applies additional checks to unreliable author metadata.

\smallskip
\noindent \textbf{Field-Level Comparison.}
Each field receives one of three outcomes, corresponding to agreement,
contradiction, and missing evidence in Section~\ref{sec:source-verification}:
\textsc{match}, \textsc{conflict}, and \textsc{missing}.
\textsc{missing} means that the available evidence is insufficient
to establish either agreement or contradiction. This can occur because
a field is absent or because the comparison falls into an ambiguous range.

Titles are compared by word overlap after normalization.
Let $A$ and $B$ be the token sets of the two titles.
We use two measures:

\[
\text{overlap}
=
\frac{|A \cap B|}{\max(|A|,|B|)},
\]

\[
\text{containment}
=
\frac{|A \cap B|}{\min(|A|,|B|)}.
\]

The first measures overall overlap, while the second captures cases where
one title is largely contained in the other.

\begin{table*}[t]
\centering
\small
\begin{tabularx}{\textwidth}{
@{}l
>{\raggedright\arraybackslash}X
>{\raggedright\arraybackslash}X
>{\raggedright\arraybackslash}X
>{\raggedright\arraybackslash}X@{}}
\toprule
Field & Comparison & \textsc{match} & \textsc{missing} & \textsc{conflict} \\
\midrule

Title
& Normalized word overlap and containment
& The titles agree after normalization, or one is a clear shortened or truncated form of the other
& The titles overlap, but not enough to establish either agreement or contradiction
& The titles are clearly different \\

Authors
& Parsed surname and given-name agreement; name order and ``et al.'' are ignored
& The cited names substantially agree with the record
& The author evidence is partial, incomplete, or otherwise inconclusive
& A reliable author list contradicts the cited authors \\

Venue
& Normalized venue names, including common aliases and abbreviations
& The venues are equivalent
& The venue is absent or the comparison is inconclusive
& The venues are clearly different \\

Year
& Publication year, allowing legitimate differences across versions
& The years are compatible with the same work
& The year is unavailable
& The years are inconsistent with the same work \\

Identifier
& Resolved DOI, arXiv identifier, or URL
& The identifier points to the same work
& The identifier is absent, incomplete, or insufficient to establish identity
& The identifier resolves to a different work \\

\bottomrule
\end{tabularx}

\caption{
Field-level evidence used in citation verification.
\textsc{missing} denotes insufficient evidence to establish either a
\textsc{match} or a \textsc{conflict}; it does not require the field itself
to be absent. These field outcomes are combined with source type and
publication-version evidence when making the final validity decision.
}
\label{tab:comparators}
\end{table*}

\smallskip
\noindent \textbf{Decision Rules.}
The field outcomes in Table~\ref{tab:comparators} are combined to determine whether the citation is supported.
Title and author evidence establishes the identity of the cited work.
Differences in venue, year, or identifier are checked against legitimate versions of the work before being treated as errors.

Retrieved candidates are examined in decreasing order of title agreement,
title similarity, year agreement, and author overlap.
For each candidate, deterministic rules first handle clear cases.
A citation is \texttt{Real} when the candidate supports the cited work without
a conflict in its identity evidence.
It is \texttt{Fake} when the evidence establishes an identity conflict, such as
incorrect authors or an identifier resolving to a different work, or when all
retrieval routes have been exhausted without finding a matching work.
Cases not settled by these rules are passed to the LLM judge.

% The deterministic rules make the final decision for 2,692 of the 3,356 citations
% (80\%).

\smallskip
\noindent \textbf{LLM Judgment for Ambiguous Cases.}
Cases not settled by the deterministic rules are passed to the LLM judge.
These include borderline title or author matches, possible differences between
publication versions, and conflicts supported by only one metadata source.

The LLM receives the parsed citation, one candidate record with its provenance,
and the field-level comparison results, with the disputed fields highlighted
(Appendix~\ref{app:prompts}). It returns \emph{error}, \emph{consistent}, or
\emph{different work}. Each judgment is sampled three times and resolved by
majority vote. If a candidate is judged to refer to a different work, it is discarded and verification proceeds to the next candidate. The only exception is when the record matches the cited title and authors, in which case we treat it as another version of the same work. We act on an \emph{error} verdict only when the title or authors conflict; discrepancies limited to the venue, year, or identifier are not treated as contradictions. This follows the same criterion used for the baselines (Appendix~\ref{app:baselines}).

Deterministic rules make the final decision for 2,692 of the 3,356 citations
(80\%), while the LLM judge resolves the remaining 664 (20\%).
Table~\ref{tab:decide-rules} summarizes the resulting decision patterns.

\begin{table*}[t]
\centering
\small

\begin{tabular}{l@{\hspace{2.6em}}c@{\hspace{2.0em}}r}
\toprule
Decision pattern & Decided by & Citations \\
\midrule

\multicolumn{3}{l}{\texttt{Real}: \textit{the citation is supported}} \\
Clear supporting evidence
    & rules & 2,378 \\
Supporting evidence requiring judgment
    & LLM & 546 \\

\midrule
\multicolumn{3}{l}{\texttt{Fake}: \textit{the citation contains contradicted fields}} \\
Clear identity contradiction
    & rules & 141 \\
Identity contradiction requiring judgment
    & LLM & 71 \\

\midrule
\multicolumn{3}{l}{\texttt{Fake}: \textit{no supported work is found}} \\
No supported record after retrieval
    & rules & 173 \\
All candidates judged to be different works
    & LLM & 47 \\

\bottomrule
\end{tabular}

\caption{
Decision patterns in evidence verification on the 3,356 CiteAudit citations
with GPT-5.6 Luna. Deterministic rules make the final decision for 2,692
citations, while LLM judgment resolves the remaining 664.
}
\label{tab:decide-rules}
\end{table*}

\smallskip
\noindent \textbf{Author Safeguards.}
Author lists in aggregated metadata are often incomplete. We therefore do not treat
an author mismatch as a conflict unless it is supported by more reliable evidence.
When a mismatch appears only in an aggregator record, we check a primary source such
as arXiv or Crossref when available. Extra cited authors count as a conflict only when
two independent records agree that those names are absent.

\subsection{Source Resolution}
\label{app:impl-resolution}

\noindent \textbf{Candidate Construction.}
$\mathsf{R}_{\theta}$ first groups records that refer to the same underlying work.
Two records are merged when their normalized titles agree and they also share a DOI,
an arXiv identifier, or the same first author. Metadata from records in the same group is consolidated: missing venue or year
information can be filled from another record, and a truncated author list can be
replaced by a longer compatible one.

\smallskip
\noindent \textbf{Candidate Support.}
A candidate is treated as a possible source only when the retrieved evidence
provides enough identity information to tie it to the citation. Concretely, we
require: (1) substantial overlap between the candidate's title and the cited title; (2) a \textsc{match} on either the
title or the authors; (3) a record of the work itself rather than a
reference-list-only entry; and (4) enough bibliographic metadata to identify the record, such as an
identifier.

Once a candidate is supported, conflicting fields are treated as discrepancies
to correct rather than reasons to discard the work. For example, a citation with
the wrong authors can still resolve to a candidate whose title and other evidence
identify the intended work.

\smallskip
\noindent \textbf{Targeted Search and Output.}
If no candidate is supported, $\mathsf{R}_{\theta}$ performs one targeted follow-up
search using the revised queries from Appendix~\ref{app:impl-discovery}, with at
most one SerpApi search and no additional page fetch. The returned records are added
to the evidence and evaluated under the same support criteria.

If exactly one work is supported, $\mathsf{R}_{\theta}$ returns its corrected
citation, with title, authors, venue, year, and identifiers taken from the retrieved
records. Otherwise, it returns \texttt{None}.
% The follow-up search is used only for
% source resolution and does not change the validity decision made by
% $\mathsf{V}_{\theta}$.

\subsection{Baseline Implementations}
\label{app:baselines}

\noindent \textbf{Direct Prompting.}
The backbone receives only the citation string and has no retrieval access.
It directly predicts one of three outcomes: the citation is correct; the citation
is incorrect but refers to an identifiable work, in which case it returns the
work's title, authors, venue, and year; or the citation is incorrect and no unique
source can be identified.

\smallskip
\noindent \textbf{CiteAudit Rerun.}
We rerun the released CiteAudit pipeline with the same backbone used in each
comparison block, replacing its original Gemini~3 Flash extractor and judge
while keeping the released prompts unchanged. The only other modification is
increasing the maximum output length from 300 to 8,192 tokens to accommodate
reasoning-model responses.

\smallskip
\noindent \textbf{Web Search.}
This baseline uses the same source-resolution prompt as Direct Prompting, but allows the
model to issue web search before answering. GPT-5.6 Luna uses OpenAI's hosted
web-search tool \footnote{\url{https://developers.openai.com/api/docs/guides/tools-web-search}}, while OpenResearcher-30B-A3B runs in its
released agent framework with the authors' browser tools and Serper backend\footnote{\url{https://serper.dev/}}
\citep{li2026openresearcher}.

\section{Evaluation Details}
\label{app:eval}

\noindent \textbf{End-to-End Source Resolution.}
All systems receive only the citation string and are not told whether it is
hallucinated. A prediction succeeds only if the system first identifies the
citation as \texttt{Fake} and then either returns the correct source for a
recoverable citation or returns \texttt{None} for an unrecoverable one.
For \method, these outputs are given by $\mathsf{V}_{\theta}$ and
$\mathsf{R}_{\theta}$, respectively. Baseline outputs are mapped to the same
two decisions. A citation predicted \texttt{Real} therefore counts as a failure
on \dataset.

\smallskip
\noindent \textbf{Matching Predicted Sources to Gold Records.}
Returned sources are compared with the gold record field by field after
normalization. Deterministic rules handle clear cases: titles must agree after
normalization; author lists are matched by name while allowing initials and
``et al.''; venues may match by normalized name, containment, acronym, or a
preprint--published version difference; and publication years may differ by at
most one year. Fields missing from the gold record are not scored.

Fields that cannot be resolved by these rules are judged by GPT-6 Sol~\citep{openai2026gpt6sol}.
The judge sees only the predicted and gold metadata, not the original citation
or the system that produced the prediction. Each ambiguous field is judged up
to three times, and the majority answer is used. As a sanity check, the judge
agreed with the deterministic decision on all 100 randomly sampled fields
already resolved by the rules.

\section{Instruction Templates}
\label{app:prompts}
\raggedbottom  % columns end short instead of stretching the gaps between boxes

% This appendix presents the instruction templates used for all LLM-based steps in our experiments: the
% three modules of \method (Appendix~\ref{app:prompts-method}) and the judge used in evaluation (Appendix~\ref{app:prompts-eval}).
% Each template gives the task, inputs, decision criteria, and output format of the corresponding prompt.

This appendix presents the instruction templates used for all LLM-based steps in our experiments: the three modules of \method (Appendix~\ref{app:prompts-method}), the ablation and the baselines (Appendix~\ref{app:prompts-baselines}), and the judge used in evaluation (Appendix~\ref{app:prompts-eval}). Each template gives the task, inputs, decision criteria, and output format of the corresponding prompt.

\subsection{\method}
\label{app:prompts-method}

\begin{promptbox}{Instruction for Field Extraction}
\textbf{Task:} Extract structured metadata from an academic citation.

\smallskip
\textbf{Input:} Citation: \texttt{\{citation\}}

\smallskip
\textbf{What to do:} Return the authors as they appear, the title of the work, the venue (journal,
conference, or publisher; \texttt{N/A} if none), and the publication year (\texttt{N/A} if none). Keep
the fields as given when they are reliable, and fix obvious OCR or formatting issues.

\textbf{Output JSON only:}

{\ttfamily\fontsize{7.6}{9.4}\selectfont\raggedright
\{\pstr{"authors"}: \pstr{"..."}, \pstr{"title"}: \pstr{"..."}, \pstr{"venue"}: \pstr{"..."}, \pstr{"year"}: \pstr{"..."}\}\par}
\end{promptbox}

\begin{promptbox}{Instruction for Source-Type Classification}
\textbf{Task:} Decide what kind of work a reference string cites.

\smallskip
\textbf{Input:} Reference: \texttt{\{citation\}}

\smallskip
\textbf{What to do:} Pick exactly one type.
\begin{itemize}[leftmargin=1.2em, itemsep=0pt, topsep=1pt]
\item \emph{paper}: a journal, conference, workshop, preprint, or working paper
\item \emph{book}: a monograph, textbook, handbook, dictionary, or edited volume
\item \emph{report}: a thesis, standard, technical report, manual, or institutional document
\item \emph{web}: a web page, such as a blog post, news story, model card, or announcement
\item \emph{dataset}: a dataset, corpus, or benchmark distribution
\item \emph{repo}: a software repository or package
\item \emph{unknown}: the reference string does not provide enough information to determine the type;
such citations follow the paper route
\end{itemize}

\smallskip
\textbf{Guidelines:} Judge the form of the work from how the reference is written. Do not judge whether
the work exists; that is decided elsewhere.

\textbf{Output JSON only:}

{\ttfamily\fontsize{7.6}{9.4}\selectfont\raggedright
\{\pstr{"type"}: \pstr{"paper | book | report | web | dataset | repo | unknown"},\\
\ \pstr{"reason"}: \pstr{"a few words"}\}\par}
\end{promptbox}

\begin{promptbox}{Instruction for Evidence Verification}
\textbf{Task:} Decide whether a citation misrepresents a retrieved record of the cited work, for the
fields that the deterministic checks could not settle.

\smallskip
\textbf{Inputs:}
\begin{itemize}[leftmargin=1.2em, itemsep=0pt, topsep=1pt, before=\raggedright]
\item Citation: \texttt{\{title\}}, \texttt{\{authors\}}, \texttt{\{year\}}, \texttt{\{venue\}}, \texttt{\{identifier\}}
\item Retrieved record from \texttt{\{source\}}: the same five fields
\item Field comparison: the outcome and score of each field, and the unsettled \texttt{\{fields\}}
\end{itemize}

\smallskip
\textbf{What to do:}
\begin{enumerate}[leftmargin=1.4em, itemsep=0pt, topsep=1pt]
\item Decide whether the citation misrepresents this work.
\item Count as an error when the citation attributes the work to authors who did not write it, gives a
title that does not describe this work, or assigns a venue or year in which no version of the work
appeared.
\item Do not count as an error: a truncated author list or ``et al.''; initials, name order, or other
variants of an author's name; a preprint cited for its published version or the reverse; an abbreviated
or alternative venue name; a year that differs between versions; another edition of a book; missing
peripheral metadata.
\item If the retrieved record is a different work altogether, answer \texttt{different\_work}. Such a
record should not be used to judge the citation.
\end{enumerate}

\textbf{Output JSON only:}

{\ttfamily\fontsize{7.6}{9.4}\selectfont\raggedright
\{\pstr{"verdict"}: \pstr{"error | consistent | different\_work"},\\
\ \pstr{"offending\_fields"}: [\pstr{"..."}],\\
\ \pstr{"reason"}: \pstr{"one sentence"}\}\par}
\end{promptbox}

\begin{promptbox}{Field Comparison Shown to the LLM}
The field comparison lists one entry per field: the outcome shown to the LLM (\texttt{agrees},
\texttt{CONFLICTS}, \texttt{undecided}, or \texttt{AMBIGUOUS} when between thresholds) and the computed
scores. \texttt{undecided} and \texttt{AMBIGUOUS} are grouped as \textsc{missing} in
Table~\ref{tab:comparators}. Two examples from the GPT-5.6 Luna run follow.

\smallskip
\textbf{(a) Title between thresholds.} Cited: \emph{Industry 4.0, smart manufacturing, and industrial
engineering: Challenges and opportunities} (2024). Record (Crossref): \emph{Industry 4.0, Smart
Manufacturing, and Industrial Engineering}, same three authors, 2024.

{\ttfamily\fontsize{7.2}{8.8}\selectfont\raggedright
title\ \ \ \ \ \ AMBIGUOUS (between thresholds)\\
\ \ \ \ \ \ \ \ \ \ \ \ similarity 0.80, containment 1.00\\
authors\ \ \ \ agrees\\
\ \ \ \ \ \ \ \ \ \ \ \ 1.00 of cited authors present\\
year\ \ \ \ \ \ \ agrees\\
venue\ \ \ \ \ \ undecided\ \ score 0.00\\
identifier\ undecided\par}
\smallskip
The LLM returned \emph{consistent}: the citation adds a subtitle but refers to the same work.

\smallskip
\textbf{(b) Exact title, authors in conflict on a single record.} Cited: \emph{ProtST: Multi-modality
learning of protein sequences and biomedical texts}, arXiv:2301.12040, 2023, with three authors. Record
(arXiv): the same title, identifier, and year, with four different authors.

{\ttfamily\fontsize{7.2}{8.8}\selectfont\raggedright
title\ \ \ \ \ \ agrees\\
\ \ \ \ \ \ \ \ \ \ \ \ similarity 1.00, containment 1.00\\
authors\ \ \ \ CONFLICTS\\
\ \ \ \ \ \ \ \ \ \ \ \ 0.00 of cited authors present;\\
\ \ \ \ \ \ \ \ \ \ \ \ NOT ON THE RECORD'S COMPLETE AUTHOR LIST:\\
\ \ \ \ \ \ \ \ \ \ \ \ \{the three cited names\}\\
year\ \ \ \ \ \ \ agrees\\
venue\ \ \ \ \ \ agrees\ \ \ \ \ score 1.00\\
identifier\ agrees\par}
\smallskip
The LLM flagged the author field as an error. The arXiv record identifies the intended work.
\end{promptbox}

\subsection{Ablation and Baselines}
\label{app:prompts-baselines}

\begin{promptbox}{Instruction for Single-Call Judgment}
\textbf{Task} (\emph{w/o Chain}): Given a citation and the records retrieved for it, decide in one
call whether the citation is correct, hallucinated with its source among the records, or hallucinated
without one.

\smallskip
\textbf{Inputs:}
\begin{itemize}[leftmargin=1.2em, itemsep=0pt, topsep=1pt, before=\raggedright]
\item Citation: \texttt{\{citation\}}
\item Records: \texttt{\{records\}}, numbered, each with its source, title, authors, venue, year, and
identifiers
\end{itemize}

\smallskip
\textbf{What to do:} Choose one of three answers, using these records.
\begin{enumerate}[leftmargin=1.4em, itemsep=0pt, topsep=1pt]
\item \emph{Correct}: a real work with this title by these authors exists. Only the title and the
authors decide this; a venue or year that differs from the published record does not by itself make the
citation wrong.
\item \emph{Hallucinated, recoverable}: the citation refers to a real work but gets its title or its
authors wrong, and one of the records is that work. Give the number of that record.
\item \emph{Hallucinated, unrecoverable}: %no real work corresponds to the citation
no unique source can be identified
, or none of the
records is the work it refers to.
\end{enumerate}

\textbf{Output JSON only:}

{\ttfamily\fontsize{7.6}{9.4}\selectfont\raggedright
\{\pstr{"verdict"}: \pstr{"correct | erroneous"},\\
\ \pstr{"source"}: null or the record number,\\
\ \pstr{"reason"}: \pstr{"one sentence"}\}\par}
\end{promptbox}

\begin{promptbox}{Instruction for Direct Prompting}
\textbf{Task:} Decide whether the work described by a citation exists, without any search tool.

\smallskip
\textbf{Input:} Citation: \texttt{\{citation\}}

\smallskip
\textbf{What to do:} Answer from your own knowledge.
\begin{itemize}[leftmargin=1.2em, itemsep=0pt, topsep=1pt]
\item A citation is \texttt{Real} when the work it describes exists, with this title by these authors.
Small bibliographic slips, such as a loosely named venue, a year off by one or two, a preprint cited
under its published form, or missing page numbers, do not make it \texttt{Fake}.
\item A citation is \texttt{Fake} when no such work exists, including a plausible title that nobody
wrote and a real title attached to authors who did not write it.
\item Non-academic sources, such as datasets, software, standards, documentation pages, textbooks, and
blog posts, count as real works.
\end{itemize}

\smallskip
\textbf{Guidelines:} If unsure, express it in the confidence value and still give a single best answer.

\textbf{Output JSON only:}

{\ttfamily\fontsize{7.6}{9.4}\selectfont\raggedright
\{\pstr{"exists"}: true | false, \pstr{"confidence"}: 0.0--1.0, \pstr{"reason"}: \pstr{"one sentence"}\}\par}
\end{promptbox}

\begin{promptbox}{Instruction for Source-Resolution Baselines}
\textbf{Task:} Given one entry from a reference list, decide whether it is correct, hallucinated but
recoverable, or hallucinated and unrecoverable.

\smallskip
\textbf{Input:} Citation: \texttt{\{citation\}}

\smallskip
\textbf{What to do:} Choose one of three answers.
\begin{enumerate}[leftmargin=1.4em, itemsep=0pt, topsep=1pt]
\item \emph{Correct}: a real work with this title by these authors exists. A venue or year that differs
from the published record, such as an abbreviated venue, a preprint cited for its published version, or a
year off by one, does not by itself make the citation wrong.
\item \emph{Hallucinated, recoverable}: the citation refers to a real work but gets its title or its
authors wrong. Give the correct title, full author list, venue, and year of that work.
\item \emph{Hallucinated, unrecoverable}: %no real work corresponds to the citation.
no unique source can be identified.
\end{enumerate}

\smallskip
\textbf{Guidelines:} Direct Prompting answers from its own knowledge, and Web Search may search the web first.

\textbf{Output JSON only:}

{\ttfamily\fontsize{7.6}{9.4}\selectfont\raggedright
\{\pstr{"verdict"}: \pstr{"correct | erroneous"},\\
\ \pstr{"source"}: null or \{\pstr{"title"}: \pstr{"..."},\\
\ \ \ \ \pstr{"authors"}: [\pstr{"..."}], \pstr{"venue"}: \pstr{"..."}, \pstr{"year"}: 2020\},\\
\ \pstr{"reason"}: \pstr{"one sentence"}\}\par}
\end{promptbox}

\subsection{Evaluation}
\label{app:prompts-eval}

\begin{promptbox}{Instruction for the Evaluation Judge}
\textbf{Task:} For each requested field, decide whether the value in the predicted source is correct for
the gold work.

\smallskip
\textbf{Inputs:}
\begin{itemize}[leftmargin=1.2em, itemsep=0pt, topsep=1pt, before=\raggedright]
\item Gold record: \texttt{\{title\}}, \texttt{\{authors\}}, \texttt{\{venue\}}, \texttt{\{year\}}
\item Predicted record: the same four fields
\item Fields to judge: \texttt{\{fields\}}
\end{itemize}
The judge sees neither the citation nor the system that produced the record.

\smallskip
\textbf{What to do:} Accept a field when it is:
\begin{itemize}[leftmargin=1.2em, itemsep=0pt, topsep=1pt]
\item \emph{Title}: the title of the same work or of another version of it, allowing capitalization,
punctuation, and a dropped or added subtitle.
\item \emph{Authors}: the same people, allowing initials, transliteration, name order, and a list cut short
with ``et al.'', but not a list that adds, drops, or substitutes people.
\item \emph{Venue}: a venue where some version of the work appeared, or another name of the gold venue.
\item \emph{Year}: the year of some version of the work.
\end{itemize}

\textbf{Output JSON only:}

{\ttfamily\fontsize{7.6}{9.4}\selectfont\raggedright
\{\pstr{"<field>"}: \pstr{"ok | wrong"}, ..., \pstr{"reason"}: \pstr{"one sentence"}\}\par}
\end{promptbox}

\section{More Discussion about Experiment Results}

\subsection{Published and Rerun CiteAudit Results}
\label{app:published-citeaudit}

Our CiteAudit rerun with GPT-5.6 Luna scores below the published Gemini-3-Flash result (0.762 vs.\ 0.838 F1), mainly through lower recall (0.773 vs.\ 0.951) at similar precision (0.751 vs.\ 0.749). Since the released prompts were developed for Gemini-3-Flash, we treat the backbone-matched blocks of Table~\ref{tab:citeaudit-main} as the controlled comparison and the published result as an external reference. Against this reference, \method with GPT-5.6 Luna reaches a similar F1 (0.841) at a more conservative operating point: it flags 54 valid citations instead of 149 but misses 89 hallucinated citations instead of 23.

\subsection{Resolution Outcomes by Recoverability}

Table~\ref{tab:citeresolve-main} evaluates two decisions in source resolution: whether to propose a source, and whether the proposed source is correct. With GPT-5.6 Luna, Web Search proposes a source for more Recoverable citations than \method (85.3\% vs.\ 65.9\%), but a smaller share of its proposed sources match the gold source (90.2\% vs.\ 95.2\%). On Unrecoverable citations, where the correct output is to flag the citation and withhold a correction, \method does so for 92.4\% of cases and Web Search for only 31.0\%. Since an incorrect correction adds a new false claim to a reference list, this combination of accurate proposals and reliable abstention gives \method a higher overall task success (74.7\% vs.\ 58.4\%).

\subsection{The Open-Weight Web Search Baseline}

The low task success of Web Search with OpenResearcher-30B-A3B (5.4\%) is not caused by failures of the agent framework: the model searched for 414 of the 423 citations, and all of its answers could be parsed. Instead, its corrections often repeat the errors they should fix. For Recoverable citations, it finds the right work for 129 of the 156 citations it proposes a source for, but 70 of these proposals keep the author list given in the citation, which is the erroneous field in most Recoverable citations (Table~\ref{tab:citeresolve-stats}). For Unrecoverable citations, it proposes a source for 120 of 171, although no source can be uniquely identified. \method instead takes the corrected source, including its author list, from retrieved records rather than from the model's answer, allowing the same backbone to reach 70.7\% task success.